\documentclass[11pt,a4paper]{article}
\usepackage[hyperref]{acl2017}
\usepackage{times}
\usepackage{latexsym}
\usepackage{mathtools}
\usepackage{enumitem}
\usepackage{soul}
\usepackage{graphicx}
\usepackage{booktabs}
\usepackage{multirow}
\usepackage{url}
\usepackage{etoolbox}
\usepackage{tikz}
\usepackage{amsmath}

\newcommand\encircle[1]{%
  \tikz[baseline=(X.base)] 
    \node (X) [draw, shape=circle, inner sep=0] {\strut #1};}

\aclfinalcopy

\title{Neural-based Context Representation Learning \\ for Dialog Act Classification}

\author{Daniel Ortega \qquad Ngoc Thang Vu \\
  Institute for Natural Language Processing (IMS)\\ University of Stuttgart\\ 
  Pfaffenwaldring 5b, 70569 Stuttgart, Germany\\
  {\tt \{daniel.ortega, thang.vu\}@ims.uni-stuttgart.de}}  

\date{}

\begin{document}
\maketitle
\begin{abstract}
We explore context representation learning methods in neural-based models for dialog act classification.
We propose and compare extensively different methods which combine recurrent neural network architectures and attention mechanisms (AMs) at different context levels. 
Our experimental results on two benchmark datasets show consistent improvements compared to the models without contextual information and reveal that the most suitable AM in the architecture depends on the nature of the dataset.
\end{abstract}

%%%%%%%%%%%%%%%%%%%%%%%%%%%%%%%%%%%%%%%%%%%%%%%%%%%%%%%%%%%%%%%%%%%%%%%%%%%%%%%%%%%%%%%
\section{Introduction}

The study of spoken dialogs between two or more speakers can be approached by analyzing the dialog acts (DAs), which is the intention of the speaker at every utterance during a conversation. Table \ref{tab:dialog} shows a fragment of a conversation from the Switchboard (SwDA) dataset with DA annotation.
Automatic DA classification is an important pre-processing step in natural language understanding tasks and spoken dialog systems. This classification task has been approached using traditional statistical methods such as hidden Markov models (HMMs) \cite{Stolcke:2000}, conditional random fields (CRF) \cite{Zimmermann:CRF2009} and support vector machines (SVMs) \cite{Henderson:SVM2010}. However, recent works with deep learning (DL) techniques have brought state-of-the-art models in DA classification, such as convolutional neural networks (CNNs) \cite{RCNN:KalchbrennerB13, lee:CNN_RNN_DA_2016}, recurrent neural networks (RNNs) \cite{lee:CNN_RNN_DA_2016,DBLP:journals/corr/JiHE16} and long short-term memory (LSTM) models \cite{AttentionCNN:ShenL16}.

\setlength{\belowcaptionskip}{-5pt}
\begin{table}[ht]
\resizebox{.75\textwidth}{!}{\begin{minipage}{\textwidth}
\begin{tabular}{|l|l|}
\hline \bf Utterance & \bf Dialog act \\ \hline
     A: \it Are you a musician yourself? & Yes-no-question\\
     B: \it Uh, well, I sing. & Affirmative non-yes answer \\
     A: \it Uh-huh. & Acknowledge (Backchannel)  \\
     B: \it I don't play an instrument.	& Statement-non-opinion \\
\hline
\end{tabular}
\end{minipage}}
\caption{\label{tab:dialog} Examples from the SwDA dataset. }
\end{table}

Given an utterance in a dialog without any previous context, it is not always obvious even for human beings to find the corresponding dialog act.
In many cases, the utterances are too short so that is hard to classify them, for example the utterance \textit{'Right'} can be either an \textit{Agreement} or a \textit{Backchannel} indicating the interlocutor to go on talking, in this case the context plays a key role at disambiguating.  Therefore, using context information from the previous utterances in a dialog flow is a crucial step for improving DA classification.
Few papers in the literature have suggested to utilize context as a potential knowledge source for DA classification \cite{lee:CNN_RNN_DA_2016,AttentionCNN:ShenL16}. Recently, \citet{DAcontext:Ribeiro2015} presented an extensive analysis of the influence of context on DA recognition concluding that contextual information from preceding utterances helps to improve the classification performance. Nonetheless, such information should be differentiable from the current utterance information, otherwise, the contextual information could have a negative impact.  

Attention mechanisms (AMs) introduced by \citet{Attn_ML:Bahdanau2014} have contributed to significant improvements in many natural language processing tasks, for instance 
machine translation \cite{Attn_ML:Bahdanau2014}, 
sentence classification \cite{AttentionCNN:ShenL16} and summarization \citep{summarization:RushCW15}, 
uncertainty detection \citep{attention:Adel2017},  
speech recognition \cite{Attn_speechRec:Chorowski2015}, 
sentence pair modeling \citep{setPais:YinSXZ15}, 
question-answering \citep{QA:GolubH16}, 
document classification \citep{docClass:Yang2016} 
and entailment \citep{entailment:RocktaschelGHKB15} . AMs let the model decide what parts of the input to pay attention to according to the relevance for the task.

In this paper, we explore the use of AMs to learn the context representation, as a manner to differentiate the current utterance from its context as well as a mechanism to highlight the most relevant information, while ignoring unimportant parts for DA classification.
We propose and compare extensively different neural-based methods for context representation learning by leveraging a recurrent neural network architecture with LSTM \citep{LSTM:Hochreiter1997} or gated recurrent units (GRUs) \citep{GRU:Cho14,GRU:ChungGCB14} in combination with AMs.

\section{Model}\label{sec:model}
\begin{figure*}
  \centering
  \includegraphics[width=0.9\textwidth]{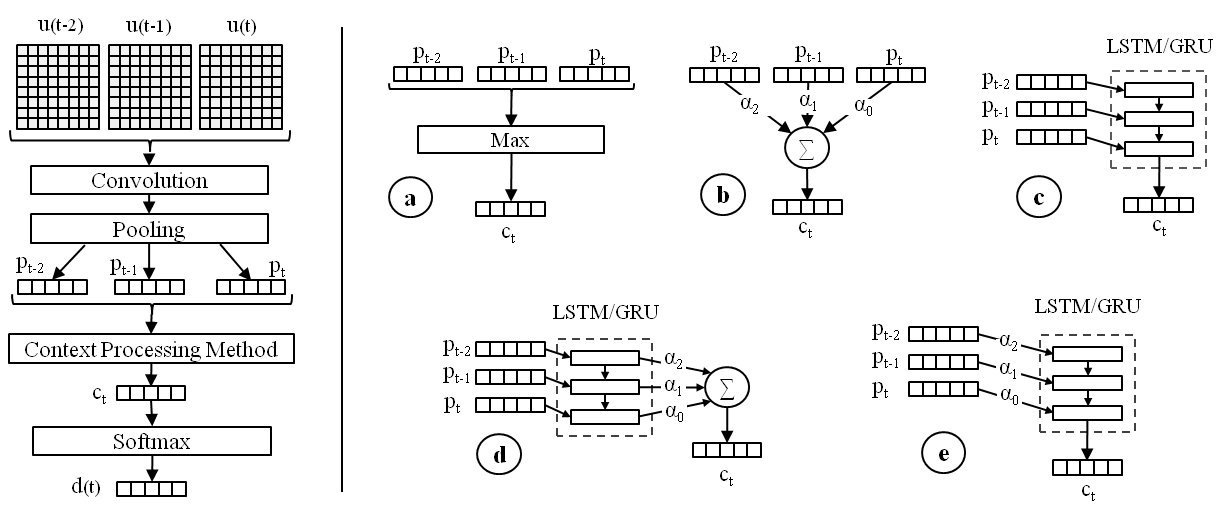}
  \caption{Model architecture for DA classification. On the left side is the overview of the model. The right site contains six neural-based methods for context representation learning.}  \label{fig:model}
\end{figure*}

The model architecture, shown on the left side of Figure \ref{fig:model}, contains two main parts: the CNN-based utterance representation and the attention mechanism for context representation learning.
Finally, the context representation is fed into a softmax layer which outputs the posterior of each predefined DA given the current dialog utterance.

\subsection{CNN-based Dialog Utterance Representation}\label{subsec:CNN}
We used CNNs for the representation of each utterance.
CNNs perform a discrete convolution on an input matrix with a set of different filters. 
For the DA classification task, the input matrix represents a dialog utterance and its context, this is $n$ previous utterances: each column of the matrix stores the word embedding of the corresponding word.
%In this work, we compare different kinds of word embeddings: randomly initialized ones and pretrained embeddings: Word2Vec \cite{WordEmbeddings:word2vec}).
We use 2D filters $f$ (with width $|f|$) spanning all embedding dimensions $d$.
This is described by the following equation:
\begin{equation}
 (w \ast f)(x,y) = \sum_{i=1}^{d}\sum_{j = -|f|/2}^{|f|/2}w(i,j) \cdot f(x-i,y-j)
\end{equation}
After convolution, a max pooling operation is applied that stores only the highest activation of each filter.
Furthermore, we apply filters with different window sizes 3-5 (multi-windows), i.e. spanning a different number of input words. 
Then, all feature maps are concatenated to one vector which represents the current utterance and its context.
\subsection{Internal Attention Mechanism}
Attention mechanisms can be applied in different sequences of input vectors, e.g.  representations of consecutive dialog utterances.
For each of the input vectors $u(t-i)$ at time step $t-i$ in a dialog and $t$ is the current time step, the attention weights $\alpha_{i} $ are computed as follows

\begin{equation}
\alpha_i = \frac{exp(f(u(t-i)))}{\sum_{0<j<m} {exp(f(u(t-j))}}
\end{equation}
where $f$ is the scoring function. 
In this work, $f$ is the linear function of the input $u(t-i)$
\begin{equation}
f(u(t-i))= W^T u(t-i)
\end{equation}
where $W$ is a trainable parameter.
The output $attentive\_u$ after the attention layer is the weighted sum of the input sequence.
\begin{equation}
attentive\_u= \sum_{i} \alpha_{i} u(t-i)
\end{equation}
Another option (\textit{order-preserved attention} as proposed in \citet{attention:Adel2017}) is to store the weighted inputs into a vector sequence $attentive\_v$ which preserves the order information.
\begin{equation}
attentive\_v= [\alpha_{0} u(t), \alpha_{1} u(t-1), ...]
\label{eq:weights}
\end{equation}
 
\subsection{Neural-based Context Modeling}
In this subsection, we present different methods, depicted on the right side of Figure \ref{fig:model}, to learn the context representation.

\paragraph{\encircle{a} Max}
We apply max-pooling on top of the dialog utterance representations which spans all the contexts and the vector dimension.
\paragraph{\encircle{b} Attention}
We apply directly attention mechanism on the dialog utterance representations.
The weighted sum of all the dialog utterances represents the context information.
\paragraph{\encircle{c} RNN}
We introduce a recurrent architecture with LSTM or GRU cells on top of the dialog utterance representations to model the relation between the context and the current utterance over time.
The output of the hidden layer of the last state is the context representation.
\paragraph{\encircle{d} RNN-Output-Attention}
Based on the previous option, we apply the attention mechanisms on the output sequence of the RNN.
The context representation is the weighted sum of all the output vectors. 
\paragraph{\encircle{e} RNN-Input-Attention}
We first apply the order-preserved attention mechanism on the dialog utterance representations to obtain a sequence of weighted inputs.
Afterwards, an RNN with LSTM or GRU cells is introduced to model the relation of the weighted context.

%%%%%%%%%%%%%%%%%%%%%%%%%%%%%%%%%%%%%%%%%%%%%%%%%%%%%%%%%%%%%%%%%%%%%%%%%%%%%%%%%%%%
\section{Experimental Setup}\label{sec:setup}

\subsection{Data}
We test our model on two DA datasets:
\begin{itemize}
\item \textbf{MRDA}: ICSI Meeting Recorder Dialog Act Corpus \cite{corpus:ICSI_orig, corpus:ICSI_annotated,techreport:ICSI_dialogAct}, a dialog corpus of \textit{multiparty meetings}. The 5-tag-set used in this work was introduced by \citet{ICSI:tagsetAng}.
\item \textbf{SwDA}: Switchboard Dialog Act Corpus \cite{Corpus:Switchboard,jurafsky97switchboard}, a dialog corpus of \textit{2-speaker conversations}. \newline
\end{itemize}
 
Train, validation and test splits on both datasets were taken as defined in \citet{lee:CNN_RNN_DA_2016}\footnote{Concerning SwDA, the data setup in \citet{lee:CNN_RNN_DA_2016} was preferred over \citet{Stolcke:2000}'s, because it was not clearly found in the latter which conversations belong to each split.}, summary statistics are shown in Table \ref{tab:datasets}. In both datasets the classes are highly unbalanced, the majority class is $59.1 $\% on MRDA and $33.7$ \% on SwDA.

\begin{table}[ht]
\centering
\resizebox{.95\textwidth}{!}{\begin{minipage}{\textwidth}
\begin{tabular}{|l|c|c|c|c|c|}
\hline \textbf{Dataset}& \textbf{C}& \textbf{$\mid$V$\mid$}& \textbf{Train}& \textbf{Validation}&\textbf{Test}\\ \hline
MRDA	&5	&12k	& 78k	&16k	&15k\\
SwDA	&43	&20k	&193k	&23k	&5k\\
\hline
\end{tabular}
\end{minipage}}
\caption{\label{tab:datasets} Data statistics: \textbf{C} is the number of classes, \textbf{$\mid$V$\mid$} is the vocabulary size and \textbf{Train}/\textbf{Validation}/\textbf{Test} are the no. of utterances.}
\end{table}

\subsection{Hyperparameters and Training}
The hyperparameters for both datasets are summarized in Table \ref{tab:hyperparams}, they were selected by varying one hyperparameter at a time while keeping the others fixed. The filter widths and feature maps were taken from the CNN architecture for sentence classification in \citet{DBLP:journals/corr/Kim14f}. 
Dropout rate of 0.5 was found to be the most effective in the range of [0-0.9]. The rectified linear unit (ReLU) was used as non-linear activation function, 1-max as pooling operation at utterance level as suggested in \citet{CNNsensitivity:Zang2015}. The only dataset specific hyperparameter is the mini-batch size: 150 and 50 for SwDA and MRDA, respectively. Word2vec \cite{WordEmbeddings:word2vec} was used for word vector representation.
Training was done for 30 epochs with averaged stochastic gradient descent \citep{ASGD:Polyak1992} over mini-batches. 
The learning rate was initialized at 0.1 and reduced 10\% every 2000 parameter updates.
We kept the word vector unchanged during training.
The context length was optimized on the development set, ranging from 1-5. 
Our best results were obtained with three context utterances for MRDA and two for SwDA. 

\begin{table}[ht]
\centering
\resizebox{.9\textwidth}{!}{\begin{minipage}{\textwidth}
\begin{tabular}{|l|c|}
\hline 
\textbf{Hyperparameter}		&\textbf{Value} \\ 
\hline
    Filter width 				&3, 4, 5 \\
    Feature maps per filter		&100 \\
    Dropout rate				&0.5 \\
 %   Learning rate				&0.1 \\
    Activation function 		&ReLU  \\
    Pooling		 				&1-max pooling per utterance  \\
   	Mini-batch size				&50 (MRDA) -- 150 (SwDA)  \\
    Word embeddings				&word2vec (dim. 300) \\
\hline
\end{tabular}
\end{minipage}}
\caption{\label{tab:hyperparams} Hyperparameters. }
\end{table}

%%%%%%%%%%%%%%%%%%%%%%%%%%%%%%%%%%%%%%%%%%%%%%%%%%%%%%%%%%%%%%%%%%%%%%%%%%%%%%%%%%%

\section{Experimental Results}\label{sec:results}

\subsection{Baseline Models}
We define two models as baseline, both are a one-layer CNN for sentence classification based on \citet{DBLP:journals/corr/Kim14f} but with an input variation: a) Baseline I: The input is a single utterance a time without any contextual information and b) Baseline II: The input is the concatenation of the current utterance and previous utterances.

\subsection{Results}
Table \ref{tab:results} summarizes the results of all the models. Results on the Baseline I and the Baseline II on both datasets show that a simple context concatenation is not enough to model the context information for this task. 
While on SwDA the accuracy improves by 1.3\%, it slightly drops on MRDA.
Other simple methods such as \emph{Max} and \emph{Attention}  do not improve the results over the baseline either.

Our results are consistently improved on both datasets after introducing RNN architecture to model the relation between the contexts.
It indicates that hierarchical structure is crucial to learn the context representation.
Attention mechanisms contribute to the overall improvements. 
On MRDA, the AM was more useful when it was applied to the inputs of the RNN, whereas on SwDA when it was applied to the outputs. 
Our intuition is that in multiparty dialogs the dependency between the utterances should be weighted before being processed by the RNN.

\begin{table}[ht]
\centering
\resizebox{0.48\textwidth}{!}{%
\begin{tabular}{|l|c|c|}
\hline
\textbf{Model} & \textbf{MRDA} & \textbf{SwDA} \\
\hline
Baseline  I		  &83.6  &71.3 \\
Baseline II   &83.5  &72.6  \\
\hline
Max & 58.5 & 48.0  \\
Attention & 83.5 & 72.4 \\
RNN (LSTM) & 83.8 & 73.1 \\
RNN (GRU) & 83.8 & 72.8 \\
RNN-Output-Attention (LSTM) & 84.1 & \textbf{73.8}  \\
RNN-Output-Attention (GRU) & 84.0 & 73.1 \\
RNN-Input-Attention (LSTM) & \textbf{84.3} & 73.3 \\
RNN-Input-Attention (GRU) & 83.6  & 73.1 \\
\hline
\end{tabular}%
}
\caption{Accuracy (\%) of baselines and models with different context processing methods.}
\label{tab:results}
\end{table}

\subsection{Impact of Context Length}\label{subsec:contextLen}
Our experiments revealed that context length plays an important role for DA classification and the best length is corpus dependent. By experimenting in the context range of 0-5 utterances, we found that the best context length for MRDA is three utterances and two for SwDA. Table \ref{tab:results_per_context} shows the results at different context lengths.

\begin{table}[ht]
\small
\centering
\begin{tabular}{|c|c|c|}
\hline 
\textbf{$n$-context}	& \textbf{MRDA}	& \textbf{SwDA} \\ 
\hline
   1 	&83.8			& 73.1\\
   2	&83.9 			&\textbf{73.8}  \\
   3 	&\textbf{84.3}	&73.5  \\
   4 	&84.0 			& 73.1 \\
   5 	&84.0 			& 72.9 \\
\hline
\end{tabular}
\caption{\label{tab:results_per_context} Comparison of accuracy (\%) on different context lengths ($n$-context, where $n$ is the number of sentences as context).}
\end{table}

\section{Comparison with Other Works}\label{sec:st_of_art}
Table \ref{tab:results_per_model} compares our results with other works.
To the best of our knowledge, \citet{lee:CNN_RNN_DA_2016} is the newest research in DA classification, which published train/validation splits and claimed to be the state-of-the-art on that setup. 
Therefore, an accurate comparison of our results can be only done with this work. 
Our model yields comparable results to the state-of-the-art on both datasets, 84.3\% against 84.6\% on MRDA and 73.8\% against 73.1\% on SwDA. 
\citet{DBLP:journals/corr/JiHE16} and \citet{RCNN:KalchbrennerB13} obtained higher accuracy on SwDA but with different setup.

\begin{table}[ht!]
\small
\centering
\begin{tabular}{|l|c|c|}
\hline 
\textbf{Model}	& \textbf{MRDA}	& \textbf{SwDA}    \\ 
\hline
    Our best model	&84.3 &73.8 \\
    CNN-FF 						&84.6 &73.1 \\
    LSTM-FF 					&84.3 &69.6 \\
    HBM  	 					&81.3 &---\\
    LV-RNN 						&---  &77.0 \\
    HCNN 						&---  &73.9 \\
    CA-LSTM						&---  &72.6 \\
    HMM 						&---  &71.0 \\
    Majority class				&59.1 &33.7 \\
\hline
\end{tabular}
\caption{\label{tab:results_per_model} Comparison of accuracy (\%).
\textit{CNN-FF} and \textit{LSTM-FF}: proposed in \citet{lee:CNN_RNN_DA_2016},
\textit{HBM}: hidden backoff model \citep{HiddenBackoff:Ji2006}. 
\textit{LV-RNN}: latent variable RNN with conditional training \citep{DBLP:journals/corr/JiHE16}.
\textit{HCNN}: hierarchical CNN \citep{RCNN:KalchbrennerB13}.
\textit{CA-LSTM}: contextual attentive LSTM \citep{AttentionCNN:ShenL16}.
\textit{HMM} \citet{Stolcke:2000}. 
}
\end{table}

\section{Conclusions}\label{sec:conclusions}
We explored different neural-based context representation learning methods for dialog act classification which combine RNN architectures with attention mechanisms at different context levels.
Our results on two benchmark datasets reveal that using RNN architecture is important to learn the context representation. 
Moreover, attention mechanisms contribute to the overall improvements, however, the place where AM should be applied depends on the nature of the dataset.

\bibliography{acl2017}

\begin{thebibliography}{}
\expandafter\ifx\csname natexlab\endcsname\relax\def\natexlab#1{#1}\fi

\bibitem[{Adel and Sch\"{u}tze(2017)}]{attention:Adel2017}
Heike Adel and Hinrich Sch\"{u}tze. 2017.
\newblock \href{http://www.aclweb.org/anthology/E17-1003}{Exploring different
  dimensions of attention for uncertainty detection}.
\newblock In {\em Proceedings of the 15th Conference of the European Chapter of
  the Association for Computational Linguistics: Volume 1, Long Papers\/}.
  Association for Computational Linguistics, Valencia, Spain, pages 22--34.
\newblock
  \href{http://www.aclweb.org/anthology/E17-1003}{http://www.aclweb.org/anthology/E17-1003}.

\bibitem[{Ang et~al.(2005)Ang, Liu, and Shriberg}]{ICSI:tagsetAng}
Jeremy Ang, Yang Liu, and Elizabeth Shriberg. 2005.
\newblock Automatic dialog act segmentation and classification in multiparty
  meetings.
\newblock In {\em in Proc. IEEE Int. Conf. on Acoustics, Speech and Signal
  Processing (ICASSP\/}. pages 1061--1064.

\bibitem[{Bahdanau et~al.(2014)Bahdanau, Cho, and
  Bengio}]{Attn_ML:Bahdanau2014}
Dzmitry Bahdanau, Kyunghyun Cho, and Yoshua Bengio. 2014.
\newblock \href{http://arxiv.org/abs/1409.0473}{Neural machine translation by
  jointly learning to align and translate}.
\newblock {\em CoRR\/} abs/1409.0473.
\newblock
  \href{http://arxiv.org/abs/1409.0473}{http://arxiv.org/abs/1409.0473}.

\bibitem[{Cho et~al.(2014)Cho, van Merrienboer, Bahdanau, and
  Bengio}]{GRU:Cho14}
KyungHyun Cho, Bart van Merrienboer, Dzmitry Bahdanau, and Yoshua Bengio. 2014.
\newblock \href{http://arxiv.org/abs/1409.1259}{On the properties of neural
  machine translation: Encoder-decoder approaches}.
\newblock {\em CoRR\/} abs/1409.1259.
\newblock
  \href{http://arxiv.org/abs/1409.1259}{http://arxiv.org/abs/1409.1259}.

\bibitem[{Chorowski et~al.(2015)Chorowski, Bahdanau, Serdyuk, Cho, and
  Bengio}]{Attn_speechRec:Chorowski2015}
Jan Chorowski, Dzmitry Bahdanau, Dmitriy Serdyuk, KyungHyun Cho, and Yoshua
  Bengio. 2015.
\newblock \href{http://arxiv.org/abs/1506.07503}{Attention-based models for
  speech recognition}.
\newblock {\em CoRR\/} abs/1506.07503.
\newblock
  \href{http://arxiv.org/abs/1506.07503}{http://arxiv.org/abs/1506.07503}.

\bibitem[{Chung et~al.(2014)Chung, G{\"{u}}l{\c{c}}ehre, Cho, and
  Bengio}]{GRU:ChungGCB14}
Junyoung Chung, {\c{C}}aglar G{\"{u}}l{\c{c}}ehre, KyungHyun Cho, and Yoshua
  Bengio. 2014.
\newblock \href{http://arxiv.org/abs/1412.3555}{Empirical evaluation of gated
  recurrent neural networks on sequence modeling}.
\newblock {\em CoRR\/} abs/1412.3555.
\newblock
  \href{http://arxiv.org/abs/1412.3555}{http://arxiv.org/abs/1412.3555}.

\bibitem[{Dhillon et~al.(2004)Dhillon, Bhagat, Carvey, and
  Shriberg}]{techreport:ICSI_dialogAct}
Rajdip Dhillon, Sonali Bhagat, Hannah Carvey, and Elizabeth Shriberg. 2004.
\newblock \href{https://goo.gl/TtLJlE}{{Meeting Recorder Project: Dialog Act
  Labeling Guide}}.
\newblock Technical report, ICSI Tech. Report.
\newblock \href{https://goo.gl/TtLJlE}{https://goo.gl/TtLJlE}.

\bibitem[{Godfrey et~al.(1992)Godfrey, Holliman, and
  McDaniel}]{Corpus:Switchboard}
John~J. Godfrey, Edward~C. Holliman, and Jane McDaniel. 1992.
\newblock \href{http://dl.acm.org/citation.cfm?id=1895550.1895693}{Switchboard:
  Telephone speech corpus for research and development}.
\newblock In {\em Proceedings of the 1992 IEEE International Conference on
  Acoustics, Speech and Signal Processing - Volume 1\/}. IEEE Computer Society,
  Washington, DC, USA, ICASSP'92, pages 517--520.
\newblock
  \href{http://dl.acm.org/citation.cfm?id=1895550.1895693}{http://dl.acm.org/citation.cfm?id=1895550.1895693}.

\bibitem[{Golub and He(2016)}]{QA:GolubH16}
David Golub and Xiaodong He. 2016.
\newblock \href{http://arxiv.org/abs/1604.00727}{Character-level question
  answering with attention}.
\newblock {\em CoRR\/} abs/1604.00727.
\newblock
  \href{http://arxiv.org/abs/1604.00727}{http://arxiv.org/abs/1604.00727}.

\bibitem[{Henderson et~al.(2012)Henderson, Gašić, Thomson, Tsiakoulis, Yu,
  and Young}]{Henderson:SVM2010}
M.~Henderson, M.~Gašić, B.~Thomson, P.~Tsiakoulis, K.~Yu, and S.~Young. 2012.
\newblock \href{https://doi.org/10.1109/SLT.2012.6424218}{Discriminative spoken
  language understanding using word confusion networks}.
\newblock In {\em 2012 IEEE Spoken Language Technology Workshop (SLT)\/}. pages
  176--181.
\newblock
  \href{https://doi.org/10.1109/SLT.2012.6424218}{https://doi.org/10.1109/SLT.2012.6424218}.

\bibitem[{Hochreiter and Schmidhuber(1997)}]{LSTM:Hochreiter1997}
Sepp Hochreiter and J\"{u}rgen Schmidhuber. 1997.
\newblock \href{https://doi.org/10.1162/neco.1997.9.8.1735}{Long short-term
  memory}.
\newblock {\em Neural Comput.\/} 9(8):1735--1780.
\newblock
  \href{https://doi.org/10.1162/neco.1997.9.8.1735}{https://doi.org/10.1162/neco.1997.9.8.1735}.

\bibitem[{Janin et~al.(2003)Janin, Baron, Edwards, Ellis, Gelbart, Morgan,
  Peskin, Pfau, Shriberg, Stolcke, and Wooters}]{corpus:ICSI_orig}
Adam Janin, Don Baron, Jane Edwards, Dan Ellis, David Gelbart, Nelson Morgan,
  Barbara Peskin, Thilo Pfau, Elizabeth Shriberg, Andreas Stolcke, and Chuck
  Wooters. 2003.
\newblock The icsi meeting corpus.
\newblock pages 364--367.

\bibitem[{Ji and Bilmes(2006)}]{HiddenBackoff:Ji2006}
Gang Ji and Jeff Bilmes. 2006.
\newblock \href{https://doi.org/10.3115/1220835.1220871}{Backoff model training
  using partially observed data: Application to dialog act tagging}.
\newblock In {\em Proceedings of the Main Conference on Human Language
  Technology Conference of the North American Chapter of the Association of
  Computational Linguistics\/}. Association for Computational Linguistics,
  Stroudsburg, PA, USA, HLT-NAACL '06, pages 280--287.
\newblock
  \href{https://doi.org/10.3115/1220835.1220871}{https://doi.org/10.3115/1220835.1220871}.

\bibitem[{Ji et~al.(2016)Ji, Haffari, and
  Eisenstein}]{DBLP:journals/corr/JiHE16}
Yangfeng Ji, Gholamreza Haffari, and Jacob Eisenstein. 2016.
\newblock \href{http://arxiv.org/abs/1603.01913}{A latent variable recurrent
  neural network for discourse relation language models}.
\newblock {\em CoRR\/} abs/1603.01913.
\newblock
  \href{http://arxiv.org/abs/1603.01913}{http://arxiv.org/abs/1603.01913}.

\bibitem[{Jurafsky et~al.(1997)Jurafsky, Shriberg, and
  Biasca}]{jurafsky97switchboard}
D.~Jurafsky, E.~Shriberg, and D.~Biasca. 1997.
\newblock {Switchboard SWBD-DAMSL shallow-discourse-function annotation coders
  manual}.
\newblock Technical Report Draft 13, University of Colorado, Institute of
  Cognitive Science.

\bibitem[{Kalchbrenner and Blunsom(2013)}]{RCNN:KalchbrennerB13}
Nal Kalchbrenner and Phil Blunsom. 2013.
\newblock \href{http://arxiv.org/abs/1306.3584}{Recurrent convolutional neural
  networks for discourse compositionality}.
\newblock {\em CoRR\/} abs/1306.3584.
\newblock
  \href{http://arxiv.org/abs/1306.3584}{http://arxiv.org/abs/1306.3584}.

\bibitem[{Kim(2014)}]{DBLP:journals/corr/Kim14f}
Yoon Kim. 2014.
\newblock \href{http://arxiv.org/abs/1408.5882}{Convolutional neural networks
  for sentence classification}.
\newblock {\em CoRR\/} abs/1408.5882.
\newblock
  \href{http://arxiv.org/abs/1408.5882}{http://arxiv.org/abs/1408.5882}.

\bibitem[{Lee and Dernoncourt(2016)}]{lee:CNN_RNN_DA_2016}
Ji~Young Lee and Franck Dernoncourt. 2016.
\newblock \href{http://arxiv.org/abs/1603.03827}{Sequential short-text
  classification with recurrent and convolutional neural networks}.
\newblock {\em CoRR\/} abs/1603.03827.
\newblock
  \href{http://arxiv.org/abs/1603.03827}{http://arxiv.org/abs/1603.03827}.

\bibitem[{Mikolov et~al.(2013)Mikolov, Sutskever, Chen, Corrado, and
  Dean}]{WordEmbeddings:word2vec}
Tomas Mikolov, Ilya Sutskever, Kai Chen, Greg Corrado, and Jeffrey Dean. 2013.
\newblock \href{http://arxiv.org/abs/1310.4546}{Distributed representations of
  words and phrases and their compositionality}.
\newblock {\em CoRR\/} abs/1310.4546.
\newblock
  \href{http://arxiv.org/abs/1310.4546}{http://arxiv.org/abs/1310.4546}.

\bibitem[{Polyak and Juditsky(1992)}]{ASGD:Polyak1992}
B.~T. Polyak and A.~B. Juditsky. 1992.
\newblock \href{https://doi.org/10.1137/0330046}{Acceleration of stochastic
  approximation by averaging}.
\newblock {\em SIAM J. Control Optim.\/} 30(4):838--855.
\newblock
  \href{https://doi.org/10.1137/0330046}{https://doi.org/10.1137/0330046}.

\bibitem[{Ribeiro et~al.(2015)Ribeiro, Ribeiro, and
  de~Matos}]{DAcontext:Ribeiro2015}
Eug{\'{e}}nio Ribeiro, Ricardo Ribeiro, and David~Martins de~Matos. 2015.
\newblock \href{http://arxiv.org/abs/1506.00839}{The influence of context on
  dialogue act recognition}.
\newblock {\em CoRR\/} abs/1506.00839.
\newblock
  \href{http://arxiv.org/abs/1506.00839}{http://arxiv.org/abs/1506.00839}.

\bibitem[{Rockt{\"{a}}schel et~al.(2015)Rockt{\"{a}}schel, Grefenstette,
  Hermann, Kocisk{\'{y}}, and Blunsom}]{entailment:RocktaschelGHKB15}
Tim Rockt{\"{a}}schel, Edward Grefenstette, Karl~Moritz Hermann, Tom{\'{a}}s
  Kocisk{\'{y}}, and Phil Blunsom. 2015.
\newblock \href{http://arxiv.org/abs/1509.06664}{Reasoning about entailment
  with neural attention}.
\newblock {\em CoRR\/} abs/1509.06664.
\newblock
  \href{http://arxiv.org/abs/1509.06664}{http://arxiv.org/abs/1509.06664}.

\bibitem[{Rush et~al.(2015)Rush, Chopra, and Weston}]{summarization:RushCW15}
Alexander~M. Rush, Sumit Chopra, and Jason Weston. 2015.
\newblock \href{http://aclweb.org/anthology/D/D15/D15-1044.pdf}{A neural
  attention model for abstractive sentence summarization}.
\newblock In Llu{\'{\i}}s M{\`{a}}rquez, Chris Callison{-}Burch, Jian Su,
  Daniele Pighin, and Yuval Marton, editors, {\em Proceedings of the 2015
  Conference on Empirical Methods in Natural Language Processing, {EMNLP} 2015,
  Lisbon, Portugal, September 17-21, 2015\/}. The Association for Computational
  Linguistics, pages 379--389.
\newblock
  \href{http://aclweb.org/anthology/D/D15/D15-1044.pdf}{http://aclweb.org/anthology/D/D15/D15-1044.pdf}.

\bibitem[{Shen and Lee(2016)}]{AttentionCNN:ShenL16}
Sheng{-}syun Shen and Hung{-}yi Lee. 2016.
\newblock \href{http://arxiv.org/abs/1604.00077}{Neural attention models for
  sequence classification: Analysis and application to key term extraction and
  dialogue act detection}.
\newblock {\em CoRR\/} abs/1604.00077.
\newblock
  \href{http://arxiv.org/abs/1604.00077}{http://arxiv.org/abs/1604.00077}.

\bibitem[{Shriberg et~al.(2004)Shriberg, Dhillon, Bhagat, Ang, and
  Carvey}]{corpus:ICSI_annotated}
Elizabeth Shriberg, Raj Dhillon, Sonali Bhagat, Jeremy Ang, and Hannah Carvey.
  2004.
\newblock \href{http://www.aclweb.org/anthology/W04-2319}{The icsi meeting
  recorder dialog act (mrda) corpus}.
\newblock In Michael Strube and Candy Sidner, editors, {\em Proceedings of the
  5th SIGdial Workshop on Discourse and Dialogue\/}. Association for
  Computational Linguistics, Cambridge, Massachusetts, USA, pages 97--100.
\newblock
  \href{http://www.aclweb.org/anthology/W04-2319}{http://www.aclweb.org/anthology/W04-2319}.

\bibitem[{Stolcke et~al.(2000)Stolcke, Coccaro, Bates, Taylor, Van Ess-Dykema,
  Ries, Shriberg, Jurafsky, Martin, and Meteer}]{Stolcke:2000}
Andreas Stolcke, Noah Coccaro, Rebecca Bates, Paul Taylor, Carol Van
  Ess-Dykema, Klaus Ries, Elizabeth Shriberg, Daniel Jurafsky, Rachel Martin,
  and Marie Meteer. 2000.
\newblock \href{https://doi.org/10.1162/089120100561737}{Dialogue act modeling
  for automatic tagging and recognition of conversational speech}.
\newblock {\em Comput. Linguist.\/} 26(3):339--373.
\newblock
  \href{https://doi.org/10.1162/089120100561737}{https://doi.org/10.1162/089120100561737}.

\bibitem[{Yang et~al.(2016)Yang, Yang, Dyer, He, Smola, and
  Hovy}]{docClass:Yang2016}
Zichao Yang, Diyi Yang, Chris Dyer, Xiaodong He, Alexander~J. Smola, and
  Eduard~H. Hovy. 2016.
\newblock Hierarchical attention networks for document classification.
\newblock In {\em HLT-NAACL\/}.

\bibitem[{Yin et~al.(2015)Yin, Sch{\"{u}}tze, Xiang, and
  Zhou}]{setPais:YinSXZ15}
Wenpeng Yin, Hinrich Sch{\"{u}}tze, Bing Xiang, and Bowen Zhou. 2015.
\newblock \href{http://arxiv.org/abs/1512.05193}{{ABCNN:} attention-based
  convolutional neural network for modeling sentence pairs}.
\newblock {\em CoRR\/} abs/1512.05193.
\newblock
  \href{http://arxiv.org/abs/1512.05193}{http://arxiv.org/abs/1512.05193}.

\bibitem[{Zhang and Wallace(2015)}]{CNNsensitivity:Zang2015}
Ye~Zhang and Byron~C. Wallace. 2015.
\newblock \href{http://arxiv.org/abs/1510.03820}{A sensitivity analysis of (and
  practitioners' guide to) convolutional neural networks for sentence
  classification}.
\newblock {\em CoRR\/} abs/1510.03820.
\newblock
  \href{http://arxiv.org/abs/1510.03820}{http://arxiv.org/abs/1510.03820}.

\bibitem[{Zimmermann(2009)}]{Zimmermann:CRF2009}
Matthias Zimmermann. 2009.
\newblock Joint segmentation and classification of dialog acts using
  conditional random fields.
\newblock In {\em INTERSPEECH\/}. pages 864--867.

\end{thebibliography}
\bibliographystyle{acl_natbib}

\end{document}